\pdfoutput=1

\documentclass[11pt]{article}

\usepackage[]{acl}

\usepackage{times}
\usepackage{latexsym}

\usepackage[T1]{fontenc}

\usepackage[utf8]{inputenc}

\usepackage{microtype}

\usepackage{inconsolata}

\usepackage{graphicx}       
\usepackage{multirow}       
\usepackage{subcaption}     
\usepackage{bold-extra}     
\usepackage{bm}             
\usepackage[hang,flushmargin]{footmisc}  
\usepackage{stfloats}  
\usepackage{amsfonts,amsmath,amssymb}
\usepackage{pifont}
\usepackage{array,booktabs,makecell,tabularx}
\usepackage{lipsum}
\usepackage{enumitem}
\usepackage{xcolor}
\usepackage{csquotes}
\usepackage{makecell}
\usepackage{changepage}

\usepackage{pifont}
\newcommand{\cmark}{\ding{51}}
\newcommand{\xmark}{\ding{55}}

\usepackage[]{todonotes}

\usepackage{tcolorbox}
\usepackage{mdframed}

\usepackage{listings}
\lstdefinelanguage{prompt}{
    frame=l,
    framerule=3pt,
    framesep=8pt,
    basicstyle=\small\ttfamily,
    commentstyle=\color{cyan},
    morecomment=[l]{//},
    moredelim=[is][\color{brown}\bfseries]{<<<}{>>>},
    moredelim=[is][\color{magenta}\bfseries]{[[[}{]]]},
    moredelim=[is][\color{orange}\bfseries]{===}{===},
    moredelim=[is][\color{blue}\bfseries]{|||}{|||},
}

\title{Identifying Factual Inconsistencies in Summaries:\\Grounding LLM Inference via Task Taxonomy}

\author{
Liyan Xu$^{1}$\thanks{Equal contributions.}\,
Zhenlin Su$^{2}$\footnotemark[1]\;
Mo Yu$^{1}$\,
Jin Xu$^{2,3}$\thanks{Co-corresponding authors.}\,
Jinho D. Choi$^{4}$\,
Jie Zhou$^{1}$\,
Fei Liu$^{4}$\footnotemark[2]\\
$^{1}$Pattern Recognition Center, WeChat AI \\
$^{2}$South China University of Technology \\ 
$^{3}$Pazhou Lab, Guangzhou \quad $^{4}$Emory University \\
\small{\texttt{liyanlxu@tencent.com} \quad \texttt{zhenlinsu75@gmail.com} \quad 
\texttt{jinxu@scut.edu.cn} \quad
\texttt{fei.liu@emory.edu}}
}

\begin{document}
\maketitle

\begin{abstract}
Factual inconsistencies pose a significant hurdle for the faithful summarization by generative models.
While a major direction to enhance inconsistency detection is to derive stronger Natural Language Inference (NLI) models, we propose an orthogonal aspect that underscores the importance of incorporating task-specific taxonomy into the inference.
To this end, we consolidate key error types of inconsistent facts in summaries, and incorporate them to facilitate both the zero-shot and supervised paradigms of LLMs.
Extensive experiments on ten datasets of five distinct domains suggest that, zero-shot LLM inference could benefit from the explicit solution space depicted by the error type taxonomy, and achieves state-of-the-art performance overall, surpassing specialized non-LLM baselines, as well as recent LLM baselines.
We further distill models that fuse the taxonomy into parameters through our designed prompt completions and supervised training strategies, efficiently substituting state-of-the-art zero-shot inference with much larger LLMs. Our data and code are publicly released at \url{https://github.com/lxucs/factax}.
\end{abstract}
\section{Introduction}
\label{sec:intro}

As abstractive summarization has been advanced significantly via generative models such as BART \cite{bart} and Large Language Models (LLMs), factual inconsistencies remain one of the key concerns for ensuring high-quality faithful summaries \cite{maynez-etal-2020-faithfulness,factcc,goyal2023news}, where certain facts from the summary are not aligned with those presented in the original document.
Previous works have studied extensively that employ various paradigms to reason inconsistencies, ranging from specialized BERT-variants \cite{bert} such as DAE \cite{dae}, QAFactEval \cite{qafacteval}, to recent LLMs equipped with general comprehension capabilities \cite{chatgpt-zs,chatgpt-star,geval}.

In particular, one outstanding direction for factual inconsistency detection is to frame it as a Natural Language Inference (NLI) problem, assessing the entailment between the document and summary \cite{nli}. Intuitively, irrelevant or inconsistent facts in the summary should reflect a low level of entailment through NLI models. Prior to LLMs, BERT-based NLI models have been successfully practiced by approaches such as SummaC \cite{summac} to identify summary inconsistencies. In this new era of LLMs, several pioneering works have shown that zero-shot prompting of LLMs is already effective with NLI-style scoring, where LLMs directly classify the summary consistency or provide a consistency score \cite{chatgpt-zs,chatgpt-star,geval}.

While it is a promising direction to keep enhancing factual inconsistency recognition by deriving stronger NLI models, such as FactCC \cite{factcc}, DocNLI \cite{docnli}, FalseSum \cite{falsesum}, AMRFact \cite{amrfact}, in this work, we propose approaches from an orthogonal aspect, which examines the incorporation of explicit solution space into the inference, such that either zero-shot LLM prompting or trained models are grounded by explicit task-specific cues, i.e. an explicit error type taxonomy.

Our motivation stems from the distinct nature between summary inconsistencies and NLI: summaries are grounded by the original document, thus leaning towards \emph{reiteration}, whereas NLI tackles a broader problem that involves \emph{extrapolation}.
Since the scope of summary inconsistency detection is roughly smaller than NLI, one can consolidate and leverage its task-specific taxonomy to rationalize a more effective inference with interpretability.

As there exist numerous annotation schemas adopted by previously introduced datasets, factual errors have been unified into a fine-grained taxonomy by \textsc{AggreFact} \cite{aggrefact}, which we consolidate upon and identify five common error types that are salient for recognizing summary inconsistencies, including \emph{Predicate Error}, \emph{Entity Error}, \emph{Circumstantial Error}, \emph{Coreference Error} and \emph{Addition Error} (Section~\ref{sec:taxonomy}), covering a wide variety of datasets (Table~\ref{tab:datasets}).
The identified error types are then utilized to anchor the inference of factual inconsistencies. 
Specifically, we examine their efficacy with LLMs in both zero-shot and supervised paradigms, and demonstrate the utility of task-specific taxonomy in complementary to the sole NLI-style classification.

For the zero-shot setting (Section~\ref{sec:zero-shot}), we craft the instruction tailored for each error type in the prompt, directing LLMs to reason specific error types according to the given guidance.
To handle long summaries, we additionally propose a window-based prompting scheme, as an effective alternative to the vanilla prompting.
For a comprehensive evaluation, our experiments are conducted on 10 datasets across five domains, including summarization on different news sources, daily or professional dialogues, official reports and narrative stories. Moreover, we employ models from OpenAI (ChatGPT, GPT-4o) along with strong open-source LLMs including Llama-3 \cite{llama2} and Mistral \cite{mistral} towards a robust conclusion.

Empirical results suggest that our proposed methods surpass all baselines, including 7 non-LLM baselines and 4 LLM baselines, showing that zero-shot LLM inference could benefit from a grounded solution space by depicting the task taxonomy in the instruction. Our proposed methods, termed Factuality with Taxonomy (\textsc{FacTax}), achieve the best overall performance across five domains; especially, \textsc{FacTax} with ChatGPT outperforms previous state-of-the-art zero-shot baseline G-Eval with GPT-4 \cite{geval,amrfact} on the \textsc{AggreFact-FtSota} benchmark \cite{aggrefact}. Our zero-shot \textsc{FacTax} methods could be seamlessly applied with stronger LLMs to harness their ongoing development; Section~\ref{sec:analysis} demonstrates that switching to the larger GPT-4o or Llama3-70B can unsurprisingly boost the improvement ``for free'', significantly surpassing previous trained models specialized for this task.

We then further seek to distill a model that fuses the task taxonomy into model parameters through supervised training. By unifying the error types of previous independently introduced datasets, we regard them jointly as training resources. Llama3-8B models are trained to learn binary decisions as well as to recognize specific error types on summaries, through our designed completions and training strategies. The resulting model outperforms previous supervised baselines, and is able to match the zero-shot \textsc{FacTax} performance with ChatGPT, effectively acting as an efficient alternative to zero-shot reasoning with much larger LLMs.

Overall, our key contributions in this work are:
\begin{itemize}[noitemsep,nolistsep,leftmargin=*]
    \item We underscore the importance of a fine-grained task taxonomy for the inference of summary inconsistencies, leading to enhanced performance and interpretability upon vanilla reasoning.
    \item We pinpoint key error types and incorporate them into our designed zero-shot prompting schemes, anchoring LLM reasoning within an explicit solution space. Experiments on diverse datasets with multiple LLMs demonstrate its efficacy.
    \item We further distill a model that rationalizes the task taxonomy into parameters through our supervised training strategies, offering SOTA performance with a smaller parameter size.
\end{itemize}

\section{Related Work}
\label{sec:related}

\begin{table*}[thbp!]
\centering
\resizebox{\textwidth}{!}{
\begin{tabular}{lcccc|ccccc}
\toprule
& Domain & Doc Len & Summ Len & \# Summ & Ent. & Pred. & Circ. & Coref. & AddE. \\
\midrule
\bf Polytope \cite{polytope} & CNN/DM & 573.2 & 64.8 & 1268 & - &  - & - & - & -\\
\bf SummEval \cite{summeval} & CNN/DM & 363.6 & 62.8 & 1698 & - &  - & - & - & -\\
\bf FRANK \cite{frank} & CNN/DM & 476.2 & 40.6 & 2246 & \cmark &  \cmark & \xmark & \xmark & \cmark\\
\bf BUMP \cite{bump} & CNN/DM & 686.4& 52.5& 1087 &\cmark &  \cmark & \cmark & \cmark & \cmark\\
\midrule
\bf CLIFF \cite{cliff} & CNN/DM \& XSum & 453.4 & 35.6 & 600 & \cmark &  \cmark & \xmark & \xmark & \cmark\\
\midrule
\bf XsumFaith \cite{xsumfaith} & XSum & 381.1 & 19.2 & 2353 & \cmark &  \cmark & \xmark & \xmark & \cmark\\
\bf QAGS/Wang'20 \cite{qags} & XSum & 324.5 & 33.3 & 474 & - &  - & - & - & -\\
\bf Goyal'21 \cite{goyal21} & XSum & 430.3 & 21.8 & 150 & \cmark &  \cmark & \xmark & \xmark & \cmark\\
\bf Cao'22 \cite{cao22} & XSum & 349.4 & 25.3 & 696 & - &  - & - & - & -\\
\midrule
\bf DiaSumFact \cite{diasumfact} & Dialogues & 187.0 & 43.7 & 475 & \cmark &  \cmark & \cmark & \cmark & \cmark\\
\bf DiaSummEval \cite{dialsummeval} & Dialogues & 109.5 & 22.6 & 474 & - &  - & - & - & -\\
\bf DiaSummFactCorr \cite{diasummfactcorr} & Dialogues & 113.1 & 20.8 & 4000 & \cmark &  \cmark & \cmark & \cmark & \cmark\\
\bf FacEval \cite{facteval} & Dialogues & 98.5 & 19.6 & 750 & \cmark &  \cmark & \cmark & \cmark & \cmark\\
\midrule
\bf GovReport \cite{govreport_fact} & Reports & 3884.5 & 397.2 & 204 & \cmark &  \cmark & \cmark & \cmark & \cmark\\
\midrule
\bf SQuALITY \cite{longeval} & Stories & 4795.7 & 376.9 & 60 & - &  - & - & - & -\\
\bottomrule
\end{tabular}}
\caption{Datasets utilized in this work with statistical details: averaged document length, summary Length, number of all available summaries; and the unified error type taxonomy described in Section~\ref{sec:taxonomy}: ``-'' means no error types originally annotated; \cmark\,and \xmark\,represent whether the corresponding error type is available after label conversion.}
\label{tab:datasets}
\end{table*}

\paragraph{Factual Inconsistency Evaluation Datasets}
Numerous datasets for evaluating factual inconsistencies in summaries have been independently introduced in recent years.
Among these, many focus on the news domain, primarily addressing CNN/DailyMail summaries \cite{cnndailymail}, such as FactCC \cite{factcc}, FRANK \cite{frank} and SummEval \cite{summeval}; others addressing XSum summaries \cite{xsum} constructed upon BBC news, such as XSumFaith \cite{xsumfaith} and DeFacto \cite{defacto}; some also addressing both, such as CLIFF \cite{cliff} and \citet{goyal21}.

Apart from news, several datasets focus on dialogue summaries, especially daily dialogues from SAMSum \cite{samsum}, such as DiaSummEval \cite{dialsummeval}, FactEval \cite{facteval}, DiaSummFactCorr \cite{diasummfactcorr}. DiaSummFact \cite{diasumfact} also assesses meeting summaries from QMSum \cite{zhong-etal-2021-qmsum}.

Recent datasets have also been proposed to address more domains, e.g. \citet{govreport_fact} evaluates factual consistency on official reports from GovReport \cite{govreport}; LongEval \cite{longeval} addresses story summaries from SQuALITY \cite{squality}.

In this work, we aim for robust evaluation across diverse domains under the same task requirement, especially for zero-shot methods that should generalize across different types of documents.

\paragraph{Non-LLM Approaches}
State-of-the-art models prior to LLMs mainly focus around two directions.
The first is to effectively leverage NLI models to assess the entailment between the document-summary pair, such as \citet{falke-etal-2019-ranking} and SummaC \cite{summac}.
Within this direction, several works focus on improving the NLI model itself through methods such as synthetic data construction \cite{factcc,docnli,falsesum,amrfact} or multitask learning \cite{alignscore,align}.
The second direction employs QA-based models, such as QuestEval \cite{questeval} and QAFactEval \cite{qafacteval}, where they generate questions regarding explicit entities in the summary, then verify upon the source document.
Apart from the two main directions, other works have also explored methods such as syntactic dependencies \cite{dae} or information extraction \cite{nan-etal-2021-entity}. Particularly, \citet{chan-etal-2023-interpretable} also emphasizes interpretable factual errors by extracting semantic frames.

\paragraph{LLM Approaches}
The capability of LLMs on detecting inconsistencies have been studied by several recent works. Most of them resolve this task in the zero-shot or few-shot prompting manner \cite{shen-etal-2023-large,chatgpt-zs,chatgpt-star,geval}. Besides, FActScore firstly recognizes atomic claims then performs LLM inference \cite{min-etal-2023-factscore}. 
Other utilization of LLMs have also been proposed, such as synthetic data generation with LLMs \cite{gekhman-etal-2023-trueteacher} or retrieval-augmented factuality detection \cite{chern2023factoolfactualitydetectiongenerative}.

\section{Task Taxonomy}
\label{sec:taxonomy}

Establishing what constitutes inconsistent facts in a summary is a fundamental aspect of this task.
In this work, we target to consolidate key error types that are salient for inconsistent fact detection in general, instead of building fine-grained complex taxonomy, for two reasons.
First, a simple taxonomy is easier to be consumed by models than a complex one, aligning with our goal of practical utilization during inference.
Second, a more fine-grained taxonomy may be of greater noises, as the annotated types from different datasets can vary significantly in their standards.

Based on the annotation schemas of previously introduced datasets and the aggregation of factual errors by \textsc{AggreFact} \cite{aggrefact}, we identify the following five salient error types:
\begin{itemize}[noitemsep,nolistsep,leftmargin=*]
    \item \emph{Predicate Error}: the semantics expressed by a predicate in the summary are not consistent with those in the source document.
    \item \emph{Entity Error}: any core arguments or attributes (e.g. subjects and objects in semantic frames) in the summary are not consistent accordingly.
    \item \emph{Circumstantial Error}: Time, duration, or the location of an event or action is not consistent.
    \item \emph{Coreference Error}: a pronoun or a reference mention in the summary cannot be resolved to refer to the correct entity.
    \item \emph{Addition Error}: the summary expresses facts or events that have no grounding sentences in the document, thus cannot be verified (unless clearly extrapolatable by common sense).
\end{itemize}

\noindent These five error types focus on the ``factuality'' aspect that reflects semantic frames not aligned with the source document, which have been partially or entirely adopted in previous datasets.
To unify labels across datasets by the above taxonomy, we conduct the following steps:

\begin{enumerate}[noitemsep,nolistsep,leftmargin=*]
    \item For datasets originally without error types annotated, no label mapping is performed.
    \item For datasets addressed by \textsc{AggreFact}, we utilize their unified labels from \textsc{AggreFact}, then perform a heuristic conversion: NP $\rightarrow$ \emph{Entity Error}; Pred $\rightarrow$ \emph{Predicate Error}; Sent $\rightarrow$ \emph{Addition Error}. For all extrinsic errors, we also mark them as \emph{Addition Error}.
    \item For datasets not included in \textsc{AggreFact}, we manually perform the label conversion per dataset (details provided in Appx.~\ref{appx:conversion}).
\end{enumerate}

\noindent Table~\ref{tab:datasets} shows the resulting conversion as well as more statistical details. We do notice that neither our adopted taxonomy nor the fine-grained one in \textsc{AggreFact} is completely free from noises, due to different annotation standards across datasets. Though, the current conversion is enough to bring positive gain as shown in Section~\ref{sec:zero-shot}.

\section{Approach: Zero-Shot Paradigm}
\label{sec:zero-shot}

\begin{figure*}[t]
\centering
\includegraphics[width=0.95\textwidth]{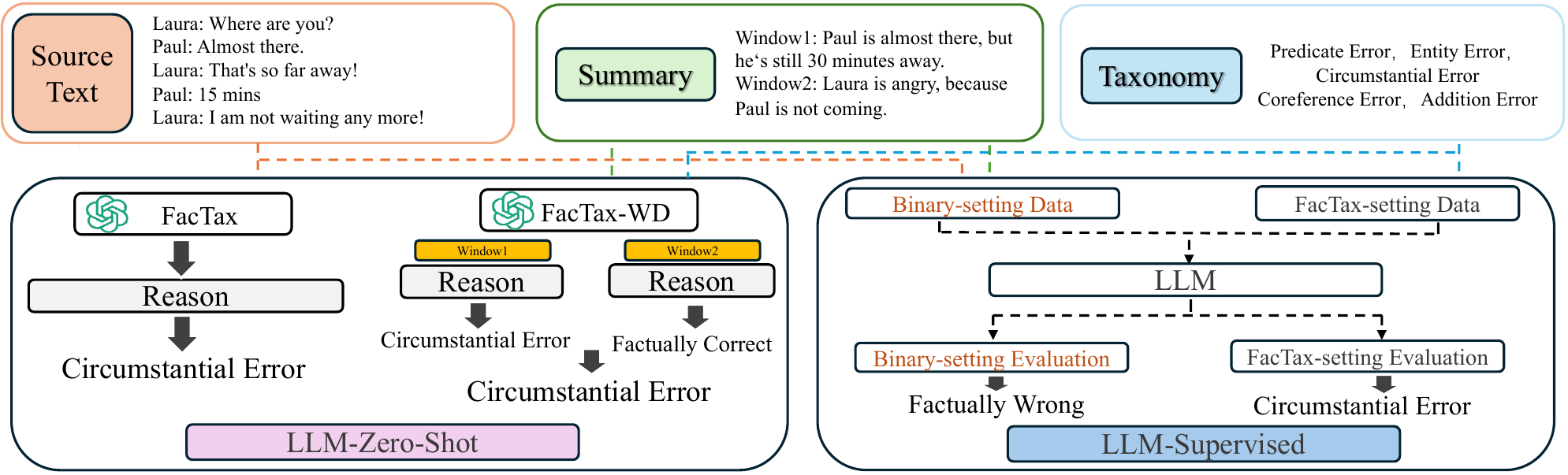}
\caption{Illustration of our proposed approaches that ground the task inference of factual inconsistency by its taxonomy (Sec.~\ref{sec:taxonomy}), via either the zero-shot paradigm (Sec.~\ref{sec:zero-shot}) or the supervised paradigm (Sec.~\ref{sec:train}) with LLMs.}
\label{fig:approach}
\end{figure*}

To incorporate the error type taxonomy, we first propose zero-shot prompting methods that leverage the general comprehension capability of LLMs, aiming to depict the explicit solution space to facilitate the zero-shot inference. As such, the proposed methods could harness the promising development of LLMs for this task, tying to greater potentials when switching to more capable LLMs.

\subsection{\textsc{FacTax}}
\label{ssec:factax}

Our first designed prompting scheme, dubbed Factuality with Taxonomy (\textsc{FacTax}), follows the standard zero-shot procedure: for a document-summary pair, we instruct a LLM to determine whether the summary is factually correct, as in previous works utilizing LLMs \cite{chatgpt-zs,chatgpt-star,geval}. For each error type, we handcraft its explanation along with an optional example, and we ask the LLM to reason in a Chain-of-Thought (CoT) style \cite{cot}: whether there are any specific error types present in the summary, instead of generating a binary decision directly. A summary is thus recognized as factually correct through a rationalization stage, whenever no specified error types are present.

The resulting zero-shot inference, to this end, is regularized by the underlying task taxonomy, so to achieve a comprehensive task reasoning. We provide our full prompt in Appx.~\ref{appx:prompts}.

\subsection{\textsc{FacTax}-WD}

Since summaries often extend beyond a single sentence, prior works adapted NLI models such as SummaC \cite{summac} conduct inference on each sentence independently, which helps mitigate degradation that may occur when inferring over long summaries.
As LLMs are susceptible to degradation over long sequences as well \cite{hsieh2024ruler}, certain errors scattered across many sentences may be overlooked by the model. Thus, we further introduce a second prompting scheme intuitively: rather than processing the entire summary at once, we divide it into separate windows that are individually processed. The final result is then aggregated across windows, such that a summary is factually correct only if each window possesses no errors. The second method is thereby termed \textsc{FacTax} by Windows (\textsc{FacTax}-WD).

After our preliminary experiments, we manually set the window size as roughly 30 words to balance between the performance and efficiency. It is worth noting that smaller window size (i.e. one sentence per window) does not necessarily lead to higher performance, as we observe that pronouns in each short window requires coreference resolution \cite{xu-choi-2020-revealing,xu-choi-2021-adapted} that could often introduce false negatives when they are inferred without its broader context.

\begin{table*}[tbp!]
\centering
\resizebox{\textwidth}{!}{
\begin{tabular}{l|ccccccccccccccccc|c}
\toprule
& \multicolumn{5}{c}{CNN/DM} && \multicolumn{5}{c}{XSum} && Dialogues && Reports && Stories \\
\cmidrule{2-6} \cmidrule{8-12} \cmidrule{14-14} \cmidrule{16-16} \cmidrule{18-18} 
& \bf Polytope & \bf SummEval & \bf Frank & \bf CLIFF & Avg.   && \bf Wang'20  & \bf CLIFF  & \bf Goyal'21  & \bf Cao'22  & Avg.  && \bf DiaSumFact  && \bf GovReport  && \bf SQuALITY & MACRO \\
\midrule
QuestEval & 17.60 & 64.90 & 62.60 & 74.00 & 70.20 && 56.00 & 61.90 & \bf 81.40 & 60.10 & 69.50 && 57.03 && 26.90 && 42.11 & 51.15  \\
QAFactEval & 32.40 & 65.20 & 54.70 & 71.60 & 67.80 && \bf 75.60 & 62.60 & 75.40 & 61.30 & 65.85 && 65.91 && 40.59 && 44.79 & 56.60 \\
SUMMAC-ZS & 97.10 & 62.20 & 57.00 & 65.60 & 64.00 && 69.80 & 59.60 & 46.60 & 49.00 & 56.40 && 58.81 && 35.19 && 15.00 & 45.88\\
\textsc{AlignScore} & 94.12 & 43.40 & 53.65 & 67.61 & 64.04 && 65.52 & 74.68 & 52.63 & 65.70 & 67.59 && 68.93 && 37.07 && 43.77 & 56.28 \\
\textsc{Align} & 91.18 & 44.92 & 55.48 & 58.30  & 69.49 && 68.09 & \bf 74.82 & 68.06 & 65.34 & 68.41 && 69.22 && 35.05 && 46.26 & 57.47\\
\sc FalseSum & - & - & - & - &  50.50 && - & - & - & - & 54.70 && - && - && - & - \\
\sc AMRFact & \bf 100.00 & \bf 80.70 & \bf 72.40 & 71.00 & \bf 72.30 && 59.50 & 66.70 & 59.10 & 64.50 & 64.10 && - && - && - & - \\
\midrule
ChatGPT-ZS & 90.19 & 79.78 & 54.82 & 65.13 & 60.03 && 71.82 & 74.01 & 63.38 & 68.82 & 69.39 && 66.85 && 41.40 && 44.63 & 56.46 \\
ChatGPT-CoT & 89.22 & 66.64 & 51.94 & 62.20 & 56.20 && 68.30 & 66.27 & 63.85 & 65.98 & 66.21 && 61.59 && 40.73 && 42.64 & 53.47 \\
ChatGPT-Star & 41.17 & 54.57 & 51.27 & 57.91 & 55.30 && 56.72 & 56.61 & 65.25& 54.89 & 55.89 && 62.86 && 35.90 && 25.62 & 47.11 \\
G-Eval & 99.02 & 48.98 & 54.18 & 56.25 & 55.04 && 51.05 & 56.61 & 53.08 & 52.36 & 51.57 && 51.73 && 15.77 && 35.86 & 41.99 \\
\midrule
\textsc{FacTax} & 78.44 & 67.43 & 62.82 & 68.71 & 68.97 && 74.06  & 70.25 & 74.08 & \bf 71.65 & \bf 72.21 && 62.76 && 40.54 && 45.93 &  58.08 \\
\textsc{FacTax-WD} & 85.31 & 72.98 & 67.09 & 70.71 & 68.92 && 71.46 & 70.81 & 68.85 & 69.49 & 69.82 && 64.15 && \bf 48.36 && \bf 48.06 & 59.94 \\
\midrule
\midrule
\textsc{FacTax} (4o) & 94.12 & 79.53 & 61.67 & \bf 79.65 & 69.60 && 74.07 & 70.21 & 73.89 & 70.05 & 71.05 && \bf 74.63 && 47.42 && 46.57 & \bf 61.85 \\
\bottomrule
\end{tabular}}
\caption{Evaluation results for the zero-shot paradigm (Section~\ref{ssec:zero-exp}). Five domains (10 datasets in total) are evaluated, where the setting for CNN/DM and XSum is kept consistent and comparable with \textsc{AggreFact-FtSota} \cite{aggrefact}. MACRO is the final evaluation metric that computes the macro-average score across each domain. \textsc{FacTax} methods are our proposed approaches that ground the zero-shot inference by the task taxonomy. All LLM-based methods are shown the averaged scores of three repeated runs for robust evaluation. The same underlying LLM (\emph{gpt-3.5-turbo-0125}) is adopted for direct comparison among LLM methods, except for \textsc{FacTax} (4o) that employs GPT-4o for demonstrating the improved performance by simply switching to stronger LLMs.}
\label{tab:result}
\vspace{-1ex}
\end{table*}

\subsection{Zero-Shot Experiments}
\label{ssec:zero-exp}

\paragraph{Datasets}
For comprehensive evaluation, we adopt diverse document types of five domains: CNN/DM, XSum, dialogues, reports, and stories. Each domain consists of one or multiple datasets from Table~\ref{tab:datasets}, with 10 datasets evaluated by the zero-shot paradigm in total. Notably, we only evaluate upon summaries generated by state-of-the-art models specified by each dataset, in coordination with \textsc{AggreFact-FtSota} \cite{aggrefact}.

For GovReport and SQuALITY, documents are long articles that can exceed the model's length limit. We follow \citet{wu2023less} that for each document, top sentences that maximize ROUGE scores towards the summary are retrieved as a condensed context (details in Appx.~\ref{appx:zero-shot}).

\paragraph{Metrics}
As in previous works, we use Balanced Accuracy for all datasets that offer classification labels. For GovReport and SQuALITY whose labels are consistency scores, we use Pearson Correlation that aligns with prior works. 
The performance of each domain is either from the standalone dataset (e.g. Dialogues), or represented by the micro average score of all datasets within this domain, aligning with \textsc{AggreFact} evaluation.
We further introduce a single metric to evaluate the overall performance, termed MACRO, which takes the macro average scores across each domain.

\paragraph{Non-LLM Methods}
We adopt strong non-LLM models as baselines, including QuestEval \cite{questeval}, QAFactEval \cite{qafacteval}, SummaC \cite{summac}, \textsc{AlignScore} \cite{alignscore} and \textsc{Align} \cite{align}. For each, we either take the evaluation scores from prior works, or run the code released by the authors on datasets not evaluated previously. Scores of \textsc{Falsesum} and \textsc{AMRFact} are from their original papers. For CNN/DM and XSum, we adopt thresholds per dataset to be consistent with previously reported numbers. More details on non-LLM baselines are provided in Appx.~\ref{appx:zero-shot}.

\paragraph{LLM Methods}
We use ChatGPT-ZS, ChatGPT-CoT \cite{chatgpt-zs}, ChatGPT-Star \cite{chatgpt-star}, and G-Eval \cite{geval} as the LLM baselines, which have achieved strong performance in \textsc{AggreFact}. For direct comparison, we use the same ChatGPT (\emph{gpt-3.5-turbo-0125}) for all LLM methods. We also vary LLMs for more insights on model comparison in Sec.~\ref{sec:analysis}.

For \textsc{FacTax}, we adapt our methods to additionally yield a score on GovReport and SQuALITY summaries to enable evaluation with their score-based labels.

\paragraph{Results}
Due to the variation of LLM generation, we run all LLM-based methods three times for robust conclusions, and show the averaged scores of each dataset in Table~\ref{tab:result}, along with evaluation results of non-LLM baselines.
Several observations from Table~\ref{tab:result} can be made as follows.

• Corroborating previous works, \textbf{LLM zero-shot inference is capable to identify factual errors directly} with decent performance, matching or exceeding strong non-LLM baselines specialized for factual inconsistency detection. Specifically, \textsc{FacTax} methods and ChatGPT-ZS achieve 56.5 - 59.9 MACRO scores, on par with 56.3 - 57.5 obtained by QAFactEval, \textsc{AlignScore} and \textsc{Align}. The lowest score of LLM baselines is 41.2, which only lags behind SUMMAC-ZS by 3.9\%.

• Comparing among LLM-based approaches, \textsc{FacTax-WD} achieves the best overall performance, surpassing the best LLM baseline ChatGPT-ZS by 3.5\%, also outperforming all non-LLM baselines. As the main difference between \textsc{FacTax} and LLM baselines is the incorporation of given task taxonomy in the prompt, the empirical result suggest that \textbf{LLM inference can indeed benefit from a grounded solution space}.

• The gap between \textsc{FacTax} and \textsc{FacTax-WD} is relatively trivial. The \textbf{window-based inference is shown effective on long summaries}, demonstrated by the significant performance raise on GovReport and SQuALITY.

• By switching to the stronger GPT-4o, \textsc{FacTax} receives significant performance boost on almost all domains (3.8\% average improvement), highlighting the flexibility, simplicity and future potentials of our proposed zero-shot methods.

\subsection{Zero-Shot Analysis}
\label{sec:analysis}

We focus on the three common domains: CNN/DM, XSum, Dialogues, and perform further analysis for more regarding insights.

\begin{table}[tbp!]
\centering
\resizebox{\columnwidth}{!}{
\begin{tabular}{l|ccc|c}
& CNN/DM & XSum & Dialogues & MACRO\\
\midrule
G-Eval (GPT-4) & \bf 69.90 & 65.80 & - & - \\
\midrule
\midrule
ChatGPT & 68.97 & \bf 72.21 & 62.76 & 67.98\\
GPT-4o & 69.60 & 71.05 & \bf 74.63 & \bf71.76\\
\midrule
Llama3-8B & 50.38 & 61.17 & 62.81 & 58.12\\
Llama3-70B & 68.56 & 71.20 &  73.08 & 70.93\\
\midrule
Mistral-7B & 51.95 & 57.80 & 58.31 & 56.02\\

\bottomrule
\end{tabular}}
\caption{Evaluation results using \textsc{FacTax} on three domains by varying LLMs. For the \textsc{AggreFact-FtSota} benchmark, \textsc{FacTax} with ChatGPT outperforms previous zero-shot G-Eval with GPT-4 (numbers reported by \citet{amrfact}).}
\label{tab:result-model}
\end{table}

\paragraph{Impact of LLMs and Sizes}
Apart from ChatGPT, we employ GPT-4o from OpenAI, as well as strong open-source LLMs including Llama3-8B/70B and Mistral-7B.
Table~\ref{tab:result-model} provides the model comparison by adopting the same \textsc{FacTax} prompts. Notably, \textsc{FacTax} with ChatGPT achieves SOTA performance on \textsc{AggreFact-FtSota} benchmark \cite{aggrefact}, outperforming G-Eval with GPT-4 \cite{amrfact}. 

Just by switching to GPT-4o, there comes a direct boost upon ChatGPT by 3+ MACRO score overall. 
There is still quite a gap of almost 10\% between the smaller 7B/8B models and the larger OpenAI models. By Table~\ref{tab:result-model}, it is evident that increasing the model size significantly improves the task reasoning, as switching Llama3-8B to 70B obtains performance gain by an impressive 12.8\%, making the zero-shot method future-proof due to the rapid LLM development.

\paragraph{Impact of Examples}
Our default \textsc{FacTax} setting grounds the inference by depicting error type definitions in the prompt, without supplying any examples. After conducting multiple rounds of experiments by adding crafted examples per type, we are not able to obtain stable improvement, as adding a few examples could lead to biases towards errors. The averaged improvement after adding examples is only 0.06 MACRO score; thus, we keep \textsc{FacTax} off examples in this work.

\begin{figure}[ht]
\centering
\includegraphics[width=0.65\columnwidth]{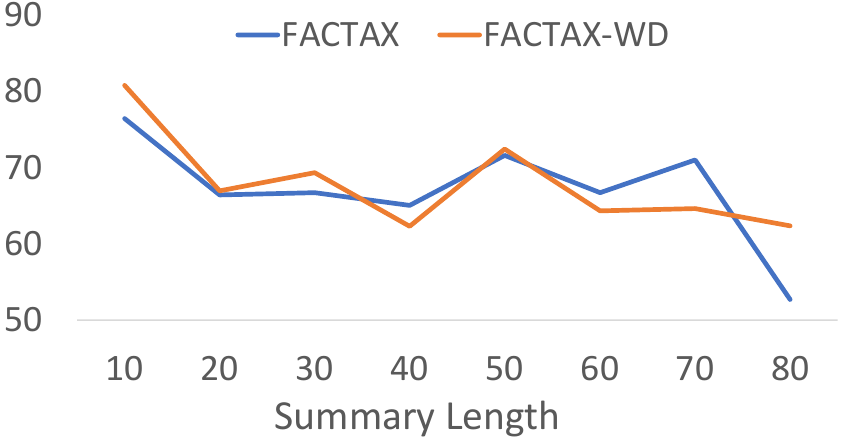}
\caption{Accuracy of \textsc{FacTax} methods for different summary lengths using ChatGPT.}
\label{fig:len1}
\end{figure}

\paragraph{Impact of Summary Lengths}
Figure~\ref{fig:len1} plots the performance curve on different lengths of summaries using ChatGPT. \textsc{FacTax-WD} is shown more robust against long summaries, due to its length-agnostic scoring mechanism. For long summaries, false positives become more often for \textsc{FacTax}, which is alleviated by the window-based inference of \textsc{FacTax-WD}. Thus, we suggest window-based inference for long summaries (e.g. > 100 tokens), while the vanilla inference should suffice for short summaries.

\section{Approach: Supervised Paradigm}
\label{sec:train}

As Section~\ref{sec:zero-shot} has demonstrated the strengths of grounding zero-shot LLM inference by its task taxonomy, we further seek to distill a model that absorbs the taxonomy into LLM parameters through supervised training.
Two advantages could come with such distillation. 
First, by learning the taxonomy from examples, the model gains real-world distribution of each error type, rather than relying on shallow comprehension through prompt instructions.
Second, it is more efficient and practical compared to zero-shot methods, avoiding the need for large model sizes and lengthy generation due to the CoT reasoning.

With above motivations, we utilize previous datasets that were proposed independently, and regard them jointly as training resources. To this end, we prepare our training and test set as follows:
\begin{itemize}[noitemsep,nolistsep,leftmargin=*]
    \item \textbf{Training Set I}: FRANK, Polytope, BUMP, CLIFF, Goyal'21, DeFacto, XSumFaith, DiaSummEval, DiaSummFactCorr, FactEval
    \item \textbf{Training Set II}: DocNLI, FalseSum
    \item \textbf{Test Set}: SummEval, Wang'20, Cao'22, DiaSumFact
\end{itemize}
\noindent Concretely, the test set is formed to cover at least one dataset per CNN/DM, XSum, and Dialogues domain. Training Set I encompasses datasets with human-annotated labels, and we use all available examples of each dataset in training, while keeping the test set only containing summaries from state-of-the-art models, such that the evaluation of the supervised paradigm is directly comparable with zero-shot results in Table~\ref{tab:result}\&\ref{tab:result-model}.

Training Set II includes two publicly released large-scale datasets constructed via synthetic data generation. 
The more recent \textsc{AMRFact} is excluded, since its data has not been released as of this writing. For efficiency, Training Set II retains randomly sampled 50k examples from DocNLI and FalseSum respectively that do not overlap with any source documents in the test set. 

The resulting training resources thereby have 16k examples in Training Set I, 100k examples in Training Set II, and 1k examples in Test Set.

\begin{table*}[tbp!]
\centering
\resizebox{0.97\textwidth}{!}{
\begin{tabular}{ll|cccccc|c}
\toprule
&& \multicolumn{1}{c}{CNN/DM} && \multicolumn{2}{c}{XSum} && Dialogues \\
\cmidrule{3-3} \cmidrule{5-6} \cmidrule{8-8}
&& \bf SummEval && \bf Wang'20 & \bf Cao'22 && \bf DiaSumFact & MACRO \\
\midrule
\multirow{2}{*}{Zero-Shot} & ChatGPT & 73.0 && 71.5 & 69.5 && 64.2 & 69.2\\
& GPT-4o & 79.5 && 74.1 & 70.1 && 74.6 & \bf 75.4\\
\midrule
\multirow{2}{*}{Zero-Shot} & Llama3-8B &  64.4 && 59.1 & 61.1 && 62.8 & 62.4\\
& Llama3-70B &77.0 && \bf 76.4 & 70.3 && 72.9 & 74.4\\
\midrule
\midrule
\multirow{4}{*}{Supervised} & \tt I-Binary + INF-Binary & 80.1 && 62.7 & \bf 72.5 && 72.6 & 73.4\\
& \tt I-Taxonomy + INF-Binary & 80.3 && 63.0 & 67.4 && 76.1 & 73.9\\
& \tt I\&II-Taxonomy + INF-Binary & 79.0 && 68.0& 70.9 && \bf 76.7 & 75.1\\
& \tt I\&II-Taxonomy + INF-Taxonomy & \bf 81.1 &&67.7& 71.8 && 75.2 & \bf 75.4\\
\bottomrule
\end{tabular}}
\caption{Evaluation results of the supervised paradigm, which are directly comparable with zero-shot results in Table~\ref{tab:result}. Llama3-8B models are trained through supervised finetuning by three training settings (Sec.~\ref{ssec:training-strategy}). For the \texttt{I\&II-Taxonomy} setting, we apply both binary inference and error type inference for the best performance.}
\label{tab:result-train}
\end{table*}

\subsection{Training Strategy}
\label{ssec:training-strategy}

With our identified task taxonomy, we unify the error types for all training examples whenever applicable according to Table~\ref{tab:datasets}.
The training is conducted through LLM supervised finetuning, where each example is converted into pairs of prompts and completions. 
To fully utilize the available resources, we design two types of prompt-completion pairs, according to if error type labels are available:

• \emph{Error Type Completion}: if a dataset has error type labels available after the label conversion, the prompt then lists error type candidates, instructing the model to generate specific error types if present any in the summary.

• \emph{Binary Completion}: for an example, the prompt can ask to directly classify if the summary is factually correct. The completion is then a binary label.

Note that for those examples with error type labels, both two types of completions can be created, which inflates training size, and also anchors different error types towards ``factually wrong''.

\subsection{Supervised Experiments}
\label{sec:train-exp}

\paragraph{Training Settings}
We employ Llama3-8B as the backbone LLM model for supervised training. Three training settings are experimented, based on different training sets and completion types:
\begin{itemize}[noitemsep,nolistsep,leftmargin=*]
    \item \texttt{I-Binary}: Training Set I that only adopts \textit{Binary Completion} for all examples.
    \item \texttt{I-Taxonomy}: Training Set I with both \textit{Binary} and \textit{Error Type Completion} when applicable.
    \item \texttt{I\&II-Taxonomy}: adding Training Set II (only \textit{Binary Completion} is applicable), in addition to all prompt-completion pairs in \texttt{I-Taxonomy}.
\end{itemize}

\noindent Particularly, the performance difference between  \texttt{I-Binary} and \texttt{I-Taxonomy} could directly reflect the impact of incorporating the task taxonomy into model parameters. \texttt{I\&II-Taxonomy} further explores the extent to which synthetic data can complement human annotations.

For the latter two settings, the INFerence of trained models is also flexible, which could either opt to determine the factual consistency directly (\texttt{INF-Binary}), or to yield fine-grained error types (\texttt{INF-Taxonomy}), according to specific types of prompts given. For \texttt{I\&II-Taxonomy}, we evaluate both inference for the best performance.

In our experiments, we adopt common hyperparameters for LLM finetuning, described in Appx.~\ref{appx:supervised}, without requiring a development set due to limited resources.
Detailed statistics of three settings are shown in Table~\ref{tab:supervised-data}. Specifically for \texttt{I\&II-Taxonomy}, we boost the ratio of \emph{Error Type Completion} to 20\% in training by adjusting the data sampling strategy, to facilitate model learning of the task taxonomy.

\begin{table}[tbp!]
\centering
\resizebox{\columnwidth}{!}{
\begin{tabular}{l|ccc|cc}
& \# Train & Length & T-Ratio & \# Test \\
\midrule
\tt I-Binary & 16393 & 648.5 & 0\% &  1033\\
\tt I-Taxonomy & 26315 & 634.2 & 37.7\% & 1033 \\
\tt I\&II-Taxonomy & 116393 & 783.3 & 20.0\% & 1033 \\
\bottomrule
\end{tabular}}
\caption{Statistics of three supervised training settings: number of prompt-completion pairs in training; averaged length of prompts; ratio of prompts with Taxonomy provided; number of prompts for evaluation.}
\label{tab:supervised-data}
\end{table}

\paragraph{Results}
Table~\ref{tab:result-train} shows the evaluation results of our supervised paradigm, along with comparison by various zero-shot results.
Unsurprisingly, trained Llama3-8B models of any settings outperform its zero-shot inference by large margins, up to 13 MACRO score. More importantly, \texttt{I\&II-Taxonomy} + \texttt{INF-Taxonomy} achieves the best performance, matching the best existing approaches by using \textsc{FacTax} with GPT-4o and Llama3-70B. Our trained model can effectively serve as \textbf{an efficient alternative to zero-shot inference by much larger LLMs}.

Comparing \texttt{I-Binary} and \texttt{I-Taxonomy}, there is an enhancement of 0.5 MACRO score by adopting \emph{Error Type Completion} in training; indeed, utilizing large-scale synthetic data brings more improvement by 1.2 MACRO score. By reasoning via error types rather than binary decisions, \texttt{I\&II-Taxonomy} receives further 0.3 gain, validating the benefit of fusing task taxonomy into model parameters.

\subsection{Supervised Analysis}
\label{ssec:supervised-analysis}

\begin{table}[tbp!]
\centering
\resizebox{\columnwidth}{!}{
\begin{tabular}{l|ccccc}
& Ent. & Pred. & Circ. & Coref. & AddE. \\
\midrule
ChatGPT  & 41.1 & 28.8 & \bf 28.2 & 21.9 &45.1 \\
GPT-4o & 60.4 & 34.4 & 18.2 & 10.0 & \bf 46.5  \\
\midrule
Llama3-8B & 32.7 & 34.2 & 23.1 & 10.4 & 35.8 \\
Llama3-70B & 58.5 & 32.6 & 27.8 & 3.4 & 43.0 \\
\midrule
\midrule
\tt I-Taxonomy & 61.1 & \bf 37.1 & 25.3 & \bf 28.9 & 21.9\\
\tt I\&II-Taxonomy & \bf 62.6 & 29.0 & 23.0 & 26.3 & 31.0 \\
\bottomrule
\end{tabular}}
\caption{F1 of five error types on DiaSumFact evaluation, with both zero-shot and supervised paradigm.}
\label{tab:error-type-f1}
\end{table}

\begin{table}[tbp!]
\centering
\resizebox{0.86\columnwidth}{!}{
\begin{tabular}{l|cccc}
& (i) & (ii) & (iii) & (iv) \\
\midrule
ChatGPT &9.7&33.4&63.4&28.5\\
GPT-4o & 9.8 &35.6&76.3& 37.6 \\
\midrule
Llama3-8B & 4.6  & 27.8 & 62.0 & 24.0 \\
Llama3-70B & 9.0  & 36.5 & 73.5 & 36.0\\
\midrule
\midrule
\tt I-Taxonomy  & 15.5 & 31.8 & \bf 85.4 & 40.4\\
\tt I\&II-Taxonomy  & \bf 15.8 & \bf 39.4& 83.8 & \bf 41.1\\
\bottomrule
\end{tabular}}
\caption{Percentage of correct error type predictions on DiaSumFact by four different criteria: i) exact match by gold error types; ii) predicted types are a subset of gold types; iii) predicted types contain one of gold types; iv) predicted types contain all gold types.}
\label{tab:case}
\end{table}

\paragraph{Fine-Grained Evaluation}
Table~\ref{tab:error-type-f1} shows the F1 score of each error type with zero-shot and supervised paradigms. Among five types, most methods suffer on \emph{Circumstantial Error} and \emph{Coreference Error}, while performing the best on \emph{Entity Error}.
The two trained models surpass zero-shot methods on three error types. However, they perform worse on \emph{Addition Error}. We attribute the degradation to different annotation standards across datasets, which may become noisy even after label unification. Nevertheless, as we have already seen improvement with the current taxonomy, future works with cleaner labels have good potentials to further boost the supervised performance.

\paragraph{Error Type Predictions}
As models often predict partially correct error types, Table~\ref{tab:case} shows the percentage of correct type predictions by four criteria, from strict to relaxed. As the results suggest, either zero-shot or supervised methods could recognize at least one gold error type on most of the factually incorrect cases, by up to 85.4\% achieved by the trained model \texttt{I-Taxonomy}. Whereas for exact match, even the trained models could only obtain 15\% accuracy. The best performance by either criterion is achieved by the supervised paradigm, as expected, since the model learns the real-world distribution from training examples.

\section{Conclusion}
\label{sec:conclusion}

We highlight the importance of task-specific taxonomy for factual inconsistency detection, where we consolidate salient error types, and incorporate them to facilitate LLM inference with both zero-shot and supervised paradigms.
Extensive experiments on ten datasets of five domains demonstrate the efficacy of depicting task taxonomy to ground the zero-shot inference, achieving state-of-the-art performance compared with respective baselines.
We further distill models that fuse the given error taxonomy into parameters through our designed training completions and strategies, effectively serving as an efficient alternative to state-of-the-art zero-shot reasoning by much larger LLMs.



\section*{Limitations}

While our study demonstrates the effective utilization of task-specific taxonomy for detecting factual inconsistencies, it is important to acknowledge certain limitations. 

First, as discussed in Section~\ref{ssec:supervised-analysis}, the unified labels after conversion can contain noises, due to the different annotation standards across previous independently introduced datasets. The resulting converted error type labels may hinder the supervised training process. Further consolidation may be conducted for a cleaner realization of the error type taxonomy.

Second, both the zero-shot paradigm and supervised paradigm may not fully capture the nuances of complex summaries. We list concrete qualitative examples in Appendix~\ref{appx:limitation} on the failed cases by LLMs.
Specifically for zero-shot paradigm, the failed cases could come from imperfect instruction following, as well as ambiguous descriptions of the task taxonomy that are not fully comprehensive. 
For the supervised paradigm, it indeed requires either human annotated examples, or synthetic data generation, which may not generalize as well as the zero-shot inference.

\section*{Acknowledgments}

We are grateful to the reviewers for their insightful comments. Liyan Xu was supported in part by the National Science Foundation grant IIS-2303678.

\bibliography{acl_latex}

\clearpage

\appendix
\section{Taxonomy Conversion}
\label{appx:conversion}

For datasets not included in \textsc{AggreFact}, we manually perform the error type conversion as follows:
\begin{itemize}[noitemsep,nolistsep,leftmargin=*]
    \item BUMP: Authors of original dataset manually edit reference summaries to constructs an unfaithful summary and classified error types into \
    Extrinsic Entity Error, Intrinsic Entity Error, Intrinsic Predicate Error, Extrinsic Circumstance Error, Intrinsic Circumstance Error, Coreference Error and Other Error. We mapped Extrinsic Entity Error and Intrinsic Entity Error to Entity Error; Extrinsic Predicate Error and Intrinsic Predicate Error to Predicate Error;  Extrinsic Circumstance Error and Intrinsic Circumstance Error to Circumstantial Error; Coreference Error to Coreference Error and Extrinsic-related Error to Addition Error. We also manually mapped five edited summaries with other types of errors to the types we have set according to our own judgment.
    \item DiaSumFact: Authors of original dataset classified error types into Ex-EntE, In-EntE, Ex-PredE, In-PredE, Ex-CirE, In-CirE, CorefE, LinkE ande Others. We mapped Ex-EntE and In-EntE to Entity Error; Ex-PredE, In-PredE and LinkE to Predicate Error; Ex-CirE and In-CirE to Circumstantial Error; CorefE to Coreference Error; Ex-Error to Addition Error and manually mapped Others base on the comment given by annotators.
    \item DiaSummFactCorr: The error types of summmaries in this dataset were classified into EntE, PredE, CircE, CorefE, LinkE, GramE, OutE and OthE. We mapped EntE to Entity Error; PredE, GramE and LinkE to Predicate Error; CircE to Circumstantial Error; CorefE to Coreference Error;OutE to Addition Error and mapped each summary with OthE manually according to our own judgment.
    \item FacEval: Authors of original dataset classified error types into Subject Object Error, Pronoun Error, Negation Error, Particulars Error, Hallucination Error and Other Error. We mapped Subject Object Error to Entity Error; Pronoun Error to Coreference Error; Negation Error to Predicate Error, Particulars Error to Circumstantial Error; Hallucination Error to Addition Error and mapped each summary with Other Error manually according to our own judgment.
    \item GovReport: Authors of original dataset classified each summary sentence's factuality based on seven types of errors: PredE, EntityE, CircE, CorefE, LinkE, OutE and GramE. We mapped EntityE to Entity Error; PredE, GramE and LinkE to Predicate Error; CircE to Circumstantial Error; CorefE to Coreference Error and OutE to Addition Error. 
    
\end{itemize}

\section{Full Prompts}
\label{appx:prompts}

We provide the full prompt for \textsc{FacTax} in Figure~\ref{fig:prompt}.

\section{Zero-Shot Experimental Settings}
\label{appx:zero-shot}

\paragraph{Long Document Alignment}
As documents in both GovReport and SQuALITY have long length of thousands of tokens, alignment is firstly performed, such that for each summary or summary window, related sentences from the document are retrieved, which will be used as a shorter context for factual error evaluation.
Though past work has proposed techniques for long context segmentation \cite{cho-etal-2022-toward}, in this work, we opt for the common approach via retrieval for simplicity.

For \textsc{FacTax}, top sentences from the document that maximize the recall of ROUGE-1 and ROUGE-2 towards the summary are retrieved until the total length reaches a certain threshold. 
These sentences are concatenated as the new context, which is shorter but has a higher information density than the original document. 

For \textsc{FacTax-WD} that operates on summary windows, $n$ important sentences are extracted independently to maximize the combined recall of ROUGE-1 and ROUGE-2 metrics in relation to the summary.
Table~\ref{tab:alignment} shows the alignment thresholds we adopted for the two datasets.

For non-LLM systems, the context length limit is usually shorter (512 for BERT models). We perform window-based inference accordingly, so that the aligned context for each summary window falls within 512 tokens. The final result is thereby aggregated through all windows.

\begin{table}[htbp!]
\centering
\resizebox{0.85\columnwidth}{!}{
\begin{tabular}{l|ccc}
& \textsc{FacTax} & \textsc{FacTax}-WD ($n$=5)  \\
\midrule
\bf GovReport & 1024 & 102.31 \\
\midrule
\bf SQuALITY  & 1024 & 28.50 \\
\bottomrule
\end{tabular}}
\caption{The maximum length of aligned context for \textsc{FacTax}, and the averaged length of aligned context per summary window for \textsc{FacTax-WD}, with $n$ being the number of sentences extracted for each summary window. For SQuALITY, some of the retrieved sentences can be quite short.}
\label{tab:alignment}
\end{table}

\paragraph{Evaluation for Baselines}
Five non-LLM baselines, QuestEval, QaFactEval, \textsc{SummaC-ZS}, \textsc{AlignScore} and \textsc{Align} produce a consistency score for each summary, which requires a threshold to convert to the classification label.
For each dataset in the \textsc{AggreFact-FtSota} test set, following \citet{aggrefact}, we tune the threshold to reach the best balanced accuracy on the corresponding \textsc{AggreFact-FtSota} validation set.

For DiaSumFact without a specific validation set, we use the \textsc{AggreFact-FtSota} validation set to tune the threshold for the non-LLM baselines.

For GovReport and SQuALITY, the evaluation metric is Pearson Correlation, thus not requiring any thresholds.

\section{Supervised Experimental Settings}
\label{appx:supervised}

We perform full finetuning of Llama3-8B with flash attention enabled on 8 Nvidia A100 GPUs (40GB memory each). Each training setting runs for 8 epochs, taking around 6 hours to finish \texttt{I-Binary} and \texttt{I-Taxonomy}, and 24 hours to finish \texttt{I\&II-Taxonomy}. We use a batch size of $1$ per GPU, learning rate of $1 \times 10^{-5}$, and cosine learning rate schedule with warm up rate $0.05$. Loss is only enabled on the completion during training.



\section{Failure Cases by LLMs}
\label{appx:limitation}

\paragraph{Failed to grasp the implied information}
For some cases, LLMs failed to grasp the implicit content in the text, leading to an error in judgment.

\begin{adjustwidth}{-0.4cm}{-0.4cm} 
\begin{quote}
\small

   \small{\textbf{[Part of the source text]}\\
    \textit{ ...\\
      Ana: You sleeping?\\
      Catherine: Not yet.\\
      Ana: Wanna go visit grandma tomorrow? I miss her.\\
      Catherine: \textbf{Yeah that would be nice} :) I'll call you when I wake up\\
      ...}}\\\\
     \small{\textbf{[Part of the summary]}\\
    \textit{Ana and Catherine are going to see their grandma tomorrow.}}\\\\
  \small{\textbf{[Human comment]}\\
    \textit{None}}\\\\
  \small{\textbf{[Error type]}\\
    \textit{Factually Correct}}\\\\
  \small{\textbf{[Response of LLM]}\\
    \textit{- Reasoning: The statement is not consistent with the source document as it is \textbf{only Ana} who expresses the desire to visit grandma, not Catherine. \\
    - Label: Entity\_Error}}

\end{quote}
\end{adjustwidth}

\paragraph{Inconsistent reasoning with humans}
For some cases, even though LLM makes the correct label prediction, its reasoning does not correspond to what humans think.

\begin{adjustwidth}{-0.4cm}{-0.4cm} 
\begin{quote}
\small

   \small{\textbf{[Part of the source text]}\\
    \textit{ ...\\
      phd d: Uh working in the cepstral domain ? So there is one guy in Grenada\\
      phd b: Yeah, in Grenada one of my friend.\\
      phd d: and another in uh, Lucent that I met at ICASSP.\\
      ...}}\\\\
     \small{\textbf{[Part of the summary]}\\
    \textit{The team said that the person in grenada was working in the cepstral domain.}}\\\\
  \small{\textbf{[Human comment]}\\
    \textit{the team -> phd d}}\\\\
  \small{\textbf{[Error type]}\\
    \textit{Entity Error}}\\\\
  \small{\textbf{[Response of LLM]}\\
    \textit{- Reasoning: The summary states that \textbf{the person in Grenada was working in the cepstral domain.} However, there is no mention in the source text about the person in Grenada working in the cepstral domain. This indicates a predicate inconsistency. \\
    - Label: Predicate\_Error}}

\end{quote}
\end{adjustwidth}
\clearpage

\begin{figure*}[t!]
    \centering
\begin{lstlisting}[language=prompt]

### Background

Now you are one of the greatest English linguist experts and Logician. Below is an instruction for evaluating the factual consistency of a given statement derived from the source document. Consistency measures whether the statement is correctly aligned with the facts presented in the source document, without contradicting or misrepresenting any information. The goal is to determine the consistency label of the statement defined as below, identifying whether the statement contains factual errors, or is factually correct towards the source document.


### Consistency Labels

i) [[[Predicate_Error]]]: a predicate from the statement (usually a verb of a semantic frame) is not consistent with the source document.
ii) [[[Entity_Error]]]: a core argument or attribute in a semantic frame (usually a subject or object) from the statement is not consistent with the source document (capitalization should not matter). 
iii) [[[Circumstantial_Error]]]: Time, duration, or the location of an event from the statement is not consistent with the source document. 
iv) [[[Coreference_Error]]]: a pronoun or a reference (e.g., this picture) from the statement is wrong, or cannot be resolved to refer to the correct entity. 
v) [[[Addition_Error]]]: the statement introduces facts that cannot be verified from the source document.
vi) [[[Factually_Correct]]]: the statement is factually correct without above factual errors. Note that it is allowed for the statement to miss important information from the source document; it is considered factually correct as long as the statement can be verified from the source document.

### Task

Please try your best to firstly reason in few sentences on whether the statement has factual errors or is factually correct, then determine the consistency label(s) from the above 6 types of labels in the end.


--- Your_Task ---

### Source Document

|||{Document}|||

### Statement

|||{Entire Summary or Summary Window}|||

### Output Format

- Reasoning: ...
- Label: ...

### Your Output

Please refer to the above instruction, return Reasoning and Label in a Markdown list, to evaluate the factual consistency of the statement.

\end{lstlisting}
    \caption{Prompt for \textsc{FacTax} described in Section~\ref{sec:zero-shot}. Slots in \textcolor{blue}{blue} refer to the input document and summary.}
    \label{fig:prompt}
\end{figure*}

\end{document}